\pdfoutput=1

\documentclass[11pt]{article}

\usepackage{emnlp2021}  

\usepackage{times}
\usepackage{latexsym}

\usepackage[T1]{fontenc}

\usepackage[utf8]{inputenc}

\usepackage{microtype}

\usepackage{graphicx}
\usepackage{subfigure}
\usepackage{multirow}
\usepackage{amsmath}

\title{Enhancing Dual-Encoders with Question and Answer Cross-Embeddings for Answer Retrieval}

\author{Yanmeng Wang$^{1}$, Jun Bai$^{2}$, Ye Wang$^{1}$, Jianfei Zhang$^{2}$, {\bf Wenge Rong}$^{2}$ \\ {\bf Zongcheng Ji$^{1}$,} {\bf Shaojun Wang}$^{1}$ \and {\bf Jing Xiao}$^{1}$
\\
         $^{1}$Ping An Technology, Beijing, China \\ $^{2}$School of Computer Science and Engineering, Beihang University, Beijing, China\\
         \{wangyanmeng219, wangye430, jizongcheng666,\\ wangshaojun851, xiaojing661\}@pingan.com.cn\\\{ba1\_jun, zhangjf, w.rong\}@buaa.edu.cn}

\begin{document}
\maketitle
\begin{abstract}
Dual-Encoders is a promising mechanism for answer retrieval in question answering (QA) systems. Currently most conventional Dual-Encoders learn the semantic representations of questions and answers merely through matching score. Researchers proposed to introduce the QA interaction features in scoring function but at the cost of low efficiency in inference stage. To keep independent encoding of questions and answers during inference stage, variational auto-encoder is further introduced to reconstruct answers (questions) from question (answer) embeddings as an auxiliary task to enhance QA interaction in representation learning in training stage. However, the needs of text generation and answer retrieval are different, which leads to hardness in training. In this work, we propose a framework to enhance the Dual-Encoders model with question answer cross-embeddings and a novel Geometry Alignment Mechanism (GAM) to align the geometry of embeddings from Dual-Encoders with that from Cross-Encoders. Extensive experimental results show that our framework significantly improves Dual-Encoders model and outperforms the state-of-the-art method on multiple answer retrieval datasets.
\end{abstract}

\section{Introduction}
Answer retrieval \cite{DBLP:conf/acl/SurdeanuCZ08} is an important mechanism in question answering (QA) systems to obtain answer candidates given a new question.
Currently, the most widely used framework for answer retrieval task is Dual-Encoders \cite{DBLP:conf/acl/SeoLKPFH19,DBLP:conf/iclr/ChangYCYK20,cer2018universal}, also known as ``Siamese Network'' \cite{triantafillou2017few,das2016together}. The Dual-Encoders model consists of two encoders to compute the embeddings of questions and answers independently, and also a predictor to estimate the relevance by a similarity score between the two embeddings.

Recently, due to the application of advanced encoding techniques, e.g., Transformer \cite{DBLP:conf/nips/VaswaniSPUJGKP17}, BERT \cite{2019/devlin/BERT}, the Dual-Encoders achieved a huge boost on the overall performance \cite{DBLP:conf/emnlp/KarpukhinOMLWEC20,DBLP:journals/corr/abs-2101-00117}. However, there remains some room to improve since the embeddings of questions and answers are encoded separately, while the cross information between questions and answers are important for answer retrieval \cite{2020/Yu/CrossVAE}.

Many efforts have been devoted in developing more powerful scoring by considering the interactions among questions and answers. For example, \citet{DBLP:conf/ijcai/XieM19a} introduced additional word-level interaction features between questions and answers for matching degree estimation. Similarly, \citet{DBLP:conf/iclr/HumeauSLW20} implemented attention mechanism to extract more information when computing matching score. Though such approaches improve the scoring mechanism, the overall efficiency derived from separate and off-line embeddings of questions and answers is sacrificed to some extent.

Therefore, it deserves discussing how to achieve better trade-off for maintaining the independent encoding in inference stage. To this end, the
Dual-VAEs \cite{2018/shen/dualVAEs} is proposed by using the question-to-question and answer-to-answer reconstruction as joint training task along with the retrieval task to improve the representation learning, which maintains the independent encoding in inference stage.
However, the embeddings produced by Dual-encoders or Dual-VAEs can still only preserve isolated information for questions or answers, while cross information between questions and answers is only learned through similarity score computed by two embeddings. Those embeddings preserving isolated semantics can lead to confusing results particularly when an answer can have multiple matched questions and vice versa, which is referred as one-to-many problem \cite{2020/Yu/CrossVAE}.

To address this challenge, \citet{2020/Yu/CrossVAE} further proposed Cross-VAEs by reconstructing answers from question embeddings and reconstructing questions from answer embeddings. 
In such way, the embeddings of questions or answers preserve the cross information from matched answers or questions and improve the performance in one-to-many cases. 
Nevertheless, both Dual-VAEs and Cross-VAEs rely on the generation sub-task to enhance the embeddings in retrieval task, while the need of text generation (the word-level joint distribution of sentences) and that of answer retrieval (the sentence-level matching distribution of QA-pairs) are different, which are suspected to conflict in joint training \cite{deudon2018learning}. It then brings an interesting question: is it feasible to exploit the cross information in retrieval task and keep the independence of sentence encoding in inference stage.

In this research we proposed a Cross-Encoders (details in section \ref{sec:cross-encoders}) as an additional guidance during Dual-Encoders training besides the similarity score. The Cross-Encoders could form comprehensive representation through cross-attention to reflect the complex relations (e.g., one-to-many) between matched questions and answers. We also developed Geometry Alignment Mechanism (details in section \ref{sec:GAM}) as the guiding way to effectively bridge the gap between Cross-Encoders and Dual-Encoders by forcing the Dual-Encoders to mimic Cross-Encoders on the geometry (i.e., semantic feature structure) in embedding space.

The contributions of this paper are in three folds: 1) Focusing on the lack of interactions in Dual-Encoders architecture, we introduce an ENhancing Dual-encoders with CROSS-Embeddings (ENDX) framework to solve this limitation, where a Cross-Encoders model is proposed to guide the training of Dual-Encoders model; 2) To achieve such enhancement in ENDX, we propose a novel Geometry Alignment Mechanism (GAM) to align the geometry of embeddings from Dual-Encoders with that from Cross-Encoders, which models the interactions between words within question and answer. This frees the Dual-Encoders from having to encode necessary information with no access to matched sentence; 
3) To validate our framework, we conduct extensive experiments and show that the proposed framework significantly improves Dual-Encoders model and outperforms the state-of-art model on multiple QA datasets.

\section{Related Work}

Traditional answer retrieval consists of two-stage pipeline including key words matching (BM25 \cite{DBLP:journals/ftir/RobertsonZ09}) to efficiently retrieve multiple relevant passages and re-ranking by neural network to select correct answers from retrieved results. But it may fall short here as the connection between answers and questions in context is not modelled directly, while the large document where the answer locates could be not highly relevant to the question \cite{DBLP:conf/acl-mrqa/AhmadCYC19}.
 
To address the problem in two-stage pipeline retrieval, there is growing interest in training end-to-end retrieval systems that can efficiently surface relevant results without an intermediate document retrieval phase \cite{DBLP:conf/emnlp/KarpukhinOMLWEC20,DBLP:conf/iclr/ChangYCYK20,DBLP:conf/acl-mrqa/AhmadCYC19,DBLP:conf/acl/SeoLKPFH19,DBLP:conf/acl/HendersonVGCBCS19}. In recent works \cite{DBLP:conf/emnlp/KarpukhinOMLWEC20,DBLP:conf/iclr/ChangYCYK20,DBLP:journals/corr/abs-2101-00117}, using dense representation learned by Dual-Encoders framework outperformed BM25 in large-scale retrieval task. Dual-Encoders can encode questions and answers independently and thus enables off-line processing to support efficient online response, but there exists a bottleneck that impedes the QA alignment for lack of interaction between questions and answers in their independent encoding.

Another popular way of sentence-level representation learning is Variational AutoEncoder (VAE). By encoding sentences into latent variables and re-constructing the same sentences from corresponding latent variables, VAE compacts the joint distribution of words in sentence into latent variable. \citet{2018/shen/dualVAEs} adopted VAE in Dual-Encoders and optimized the variational lower bound and matching loss jointly. \citet{2020/Yu/CrossVAE} proposed to reconstruct questions and answers in a crossed way to improve their interaction and allow for one-to-many projection. We do not include text reconstruction into our training goal for the difference between the need of sentence representation in reconstruction and that 
in answer retrieval.

Our proposed framework consists of a Dual-Encoders and a Cross-Encoders. The conventional Dual-Encoders provides the system with practicality in large-scale retrieval \cite{DBLP:conf/emnlp/KarpukhinOMLWEC20,DBLP:conf/iclr/ChangYCYK20,DBLP:journals/corr/abs-2101-00117}, while the Cross-Encoders has interaction between question and answer to guide the training of Dual-Encoders.

\begin{figure*}[ht]
\centering
\includegraphics[width=2\columnwidth]{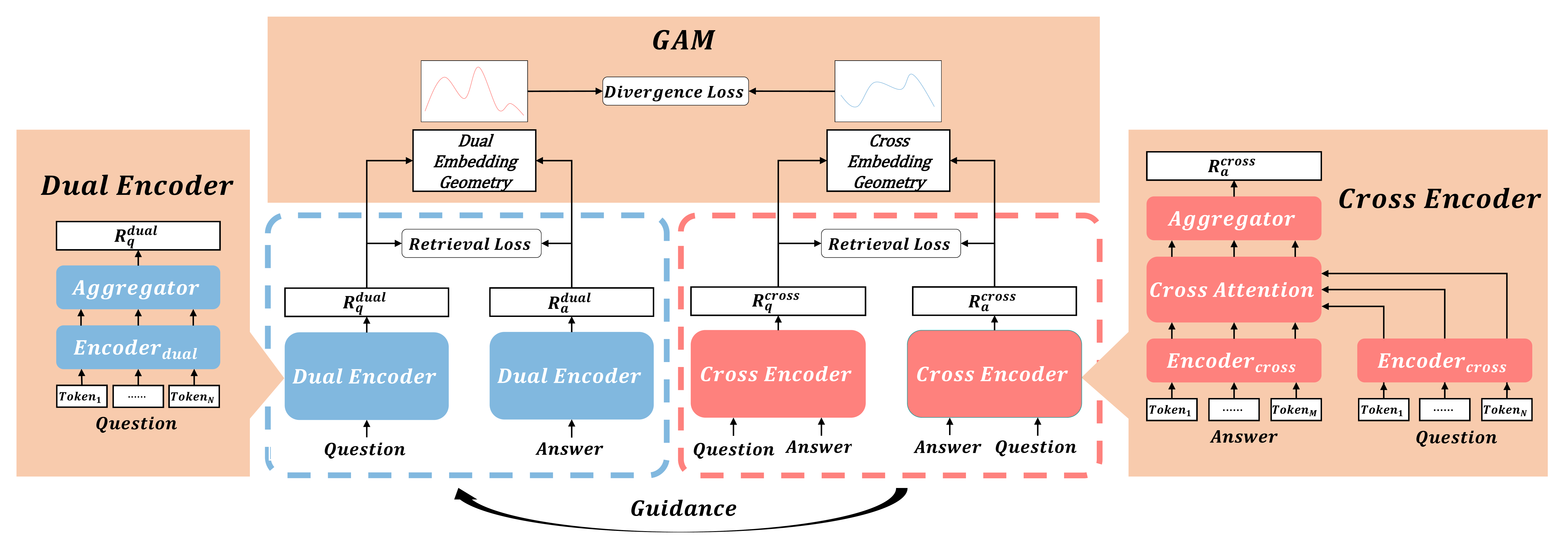}
\caption{The overview of proposed framework that enhances Dual-Encoders with cross-embeddings, Dual-Encoders (blue) and Cross-Encoders (red) are both used for training and only Dual-Encoders is used for inference.}
\label{fig:model}
\end{figure*}

\section{Methodology}

\subsection{Problem Definition}

The answer retrieval task in this work is formalized as: given a question set $S_Q$ and an answer set $S_A$, each sample could be represented as $(q, a, y)$ where $q\in S_Q$ is a question, $a\in S_A$ is a sentence-level answer, and $y$ denotes whether the answer $a$ matches the question $q$ or not. The target is to find the best-matched answer for the question $q$ and a list of candidate answers $C(q)\subset S_A$.

\subsection{Dual-Encoders}\label{sec:dual-encoders}
Our baseline model is Dual-Encoders and we refer to the sentence embedding encoded by Dual-Encoders as dual-embedding. As shown in Fig. \ref{fig:model}, the question (answer) dual-embedding $R^{dual}_q$ ($R^{dual}_a$) is processed from question (answer) text by encoder and aggregator in Dual-Encoders, where the encoder, marked as $Encoder_{dual}$ in Fig. \ref{fig:model}, can be BERT and we employ multiple hops self-attention \cite{DBLP:conf/iclr/LinFSYXZB17} as the aggregator in this work. The scoring function $f$ is defined as the inner product between the dual-embeddings of question and answer: $f(q,a) = R^{dual}_q\cdot R^{dual}_a$. Intuitively, an excellent Dual-Encoders should give high scores to matched QA pairs and low scores to mismatched QA pairs. We use in-batch negatives training strategy, which is effective for learning a Dual-Encoders model \cite{DBLP:conf/emnlp/KarpukhinOMLWEC20}. Assuming that a mini-batch has $B$ matched question-answer pairs, then the retrieval loss of a mini-batch is:
\begin{equation}\label{equ:retrieval_loss}
\mathcal{L}_{dual} = -\frac{1}{B}\sum^{B}_{i=1}\log\frac{\exp(R^{dual}_{q_i} \cdot R^{dual}_{a_i})}{\sum_{j=1}^B \exp(R^{dual}_{q_i} \cdot R^{dual}_{a_j})}
\end{equation}
where $B$ is the batch size;       $i$ and $j$ are the indexes of QA pairs in a given batch.

\subsection{Cross-Encoders}\label{sec:cross-encoders}

The cross-embeddings that involve rich question-answer interaction are obtained from the Cross-Encoders. As shown in Fig \ref{fig:model}, the Cross-Encoders gets input from both answer and question sentences. To capture precise question-answer interaction, the matched answer (question) is used to guide the encoding of question (answer). 

Let $H_q \in R^{N \times d_r}$ and $H_a \in R^{M \times d_r}$ denote the contextualized representations of words in question and answer sentences from $Encoder_{cross}$ respectively, where $N$ and $M$ are the number of words in question and answer sentences and $d_r$ is the dimension of contextualized representation. A multi-head scaled dot-production attention \cite{DBLP:conf/nips/VaswaniSPUJGKP17}, marked as $Cross\ Attention$ in Fig. \ref{fig:model}, is used to refine question (answer) contextualized representation by matched answer (question). Take the refinement of question for instance, the $i^{th}$ head is calculated as Eq. \ref{equ:head} and all heads are concatenated as Eq. \ref{equ:headcat} to obtain the answer-attended question representation $H^{'}_q$, then position-wise feed-forward networks (FFN) and layer normalization (LayerNorm) are used to further refine $H^{'}_q$ to obtain enhanced question contextualized representation $H^{cross}_q$ as Eq. \ref{equ:ffn}:
\begin{equation}\label{equ:head}
head^{i}_q = {\rm softmax}(\frac{H_{a}W^i_q(H_qW^i_k)^T}{\sqrt{d_h}})H_qW^i_v
\end{equation}
\begin{equation}\label{equ:headcat}
H^{'}_q = [head^{1}_q;...;head^{l_h}_q]W_o
\end{equation}
\begin{equation}\label{equ:ffn}
H^{cross}_q = {\rm LayerNorm}(H^{'}_q + {\rm FFN}(H^{'}_q))
\end{equation}

\noindent where $H^{cross}_q \in R^{M \times d_r}$; $l_h$ is the number of heads; $W^i_q$, $W^i_k$, $W^i_v$ and $W_o$ are learnable weights. Similarly we can obtain the enhanced answer contextualized representation $H^{cross}_a \in R^{N \times d_r}$. Multi-head attention can model word-level relationships across question and answer, and reflect the similarity between every pair of word contextualized representation across question and answer to capture the question-answer interaction and to form the comprehensive embedding of source sentence.

The sequence of $H^{cross}_q$ and $H^{cross}_a$ are then aggregated into fixed-length cross-embeddings $R^{cross}_q$ and $R^{cross}_a$, which can precisely model the relations between questions and answers. The Cross-Encoders can be trained through loss function that is defined on a mini-batch as Eq. \ref{equ:rank_loss_post}:
\begin{equation}\label{equ:rank_loss_post}
\mathcal{L}_{cross} = -\frac{1}{B}\sum^{B}_{i=1}\log\frac{\exp(R^{cross}_{q_i} \cdot R^{cross}_{a_i})}{\sum_{j=1}^B \exp(R^{cross}_{q_i} \cdot R^{cross}_{a_j})}
\end{equation}
where $B$ is the batch size; $i$ and $j$ are the indexes of the QA pairs in a given batch.

\subsection{Geometry Alignment Mechanism}\label{sec:GAM}

The dual-embeddings mechanism can save much response time through off-line processing while the cross-embeddings introduce early interaction and produce retrieved answer set with better relevance. To meet the gap between the dual-embeddings and cross-embeddings, regression is a direct way that can be easily thought of. However, this element-wise alignment in high dimensional space is too rigid for answer retrieval. 

Inspired by the geometry-preserved dimensionality reduction for pair-wise interaction modeling proposed in SNE \cite{DBLP:conf/nips/HintonR02}, we relax the element-wise alignment to the pair-wise alignment in the form of geometry, which is also proved to be crucial in representation learning 
\cite{DBLP:conf/eccv/PassalisT18}. Therefore, in this research we propose the Geometry Alignment Mechanism (GAM) to align the geometry of dual-embeddings with that of cross-embeddings, which capture the question-answer interaction. Specifically, the geometry of embeddings tells who are the neighbors of a question or an answer in the embedding space. In other words, it tells which question-answer pairs, question-question pairs or answer-answer pairs are likely to be close in the feature space.    

Since Dual-Encoders are not able to exploit the information from matched questions or answers, it might be difficult to accurately recreates the whole geometry of cross-embedding. Therefore we use the conditional probability converted from pair-wise dissimilarities to represent the geometry of data sample in feature space \cite{DBLP:conf/nips/HintonR02,van2008visualizing}. The conditional probability expresses the asymmetric probability of each datapoint $e_i$ being close to another datapoint $e_j$ in feature space as Eq. \ref{equ:cond_prob}: 

\begin{equation}\label{equ:cond_prob}
p(e_j|e_i)=\frac{\exp{(-d(e_i,e_j))}}{\sum_{k}{\exp{(-d(e_i,e_k))}}}
\end{equation}
where the $d(e_i,e_j)$ measures the dissimilarity between $e_i$ and $e_j$.

Consequently the probability of question $q_i$ being close to answer $a_j$ in feature space can be described by the conditional probability $p(a_j|q_i)$. 
To estimate such probabilities, we can use kernel density estimation (KDE) \cite{DBLP:books/wi/Scott92}, which replaces the negative dissimilarity function $-d(e_i,e_j)$ with a symmetric kernel function $K(e_i,e_j;\sigma^2)$ to model the similarity between $e_i$ and $e_j$, where $\sigma^2$ is width. The conditional probability $p(a_j|q_i)$ of cross-embeddings $p_{cross}(a_{j}|q_{i})$ and that of dual-embeddings $p_{dual}(a_{j}|q_{i})$ can be estimated using a batch of samples as Eqs. \ref{equ:cond_cross} and \ref{equ:cond_dual} consequently:
\begin{equation}\label{equ:cond_cross}
\begin{split}
&p_{cross}(a_{j}|q_{i})\\
&= \frac{\exp{(K(R^{cross}_{q_{i}},R^{cross}_{a_{j}};2\sigma^2_{c_{aq}}))}}{\sum^{B}_{k=1}{\exp{(K(R^{cross}_{q_{i}},R^{cross}_{a_{k}};2\sigma^2_{c_{aq}}))}}}
\end{split}
\end{equation}
\begin{equation}\label{equ:cond_dual}
\begin{split}
&p_{dual}(a_{j}|q_{i})\\
&=\frac{\exp{(K(R^{dual}_{q_{i}},R^{dual}_{a_{j}};2\sigma^2_{d_{aq}}))}}{\sum^{B}_{k=1}{\exp{(K(R^{dual}_{q_{i}},R^{dual}_{a_{k}};2\sigma^2_{d_{aq}}))}}}
\end{split}
\end{equation}
where $B$ is the batch size; $i$, $j$ and $k$ are the indexes of the QA pairs in a given batch.

The conditional probabilities $p(q_{j}|q_{i})$, $p(a_{j}|a_{i})$ can be estimated similarly. Since the conditional probability is asymmetric, $p(q_{j}|a_{i})$ is also needed.
One of the most natural choices of the kernel for kernel density estimation is Gaussian kernel defined as Eq. \ref{equ:Gausskernel}, while it suffers from the need of well-tuned width \cite{turlach1993bandwidth}:
\begin{equation}\label{equ:Gausskernel}
K_{Gaussian}(e_i,e_j;\sigma)=\exp{(-\frac{{\Vert e_i-e_j \Vert}^2_2}{\sigma})}
\end{equation}

To alleviate the problem of domain-dependent tuning and adapt the kernel to our scoring function, we use inner product-based similarity metric as defined in Eq. \ref{equ:Innerkernel}:
\begin{equation}\label{equ:Innerkernel}
K_{Inner}(e_i,e_j)=e_i^{T}e_j
\end{equation}

\begin{table*}
	\begin{center}
		\resizebox{2\columnwidth}{!}
		{  
			\begin{tabular}{l|c|c|c|c|c|c|c|c|c}
				\hline
				\multirow{2}*{Dataset} & \multicolumn{5}{c|}{Training} & \multicolumn{4}{c}{Test} \\
				\cline{2-10}
				~ & \#Q & \#A & \#QA pairs & \#A per Q & \#Q per A & \#Q & \#A & \#A per Q & \#Q per A \\
				\hline
				ReQA SQuAD & 87,355 & 58,934 & 87,599 & 1.00 & 1.48 & 10,539 & 7087  & 1.08 & 1.61 \\
				ReQA NQ & 104,600 & 83,153 & 107,082 & 1.03 & 1.29 & 4,177 & 5799 & 1.43 & 1.03 \\
				ReQA HotpotQA & 72,921 & 57,485 & 72,928 & 1.00 & 1.27 & 5,901 & 5745 & 1.00 & 1.03 \\
				ReQA NewsQA & 71,561 & 39,415 & 74,160 & 1.03 & 1.88 & 4,185 & 2351 & 1.01 & 1.79 \\
				\hline
			\end{tabular}
		}
	\end{center}
        \caption{Datasets statistics. 
		\#Q denotes the number of questions. \#A per Q denotes the average number of matched answers for each question, and \#Q per A denotes the average number of matched questions for each answer.}
		\label{table: DatasetsStatistics}
\end{table*}

In order that dual-embeddings of questions $q_i$ and $q_j$ can precisely model the similarity between the cross-embeddings of questions $q_i$ and $q_j$, 
the conditional probabilities $p_{dual}(q_j|q_i)$ and $p_{cross}(q_j|q_i)$ should be as close as possible. Therefore, the GAM aims to learn a dual-embeddings representation that can minimize the divergence between $p_{dual}(q_j|q_i)$ and $p_{cross}(q_j|q_i)$, $p_{dual}(a_j|a_i)$ and $p_{cross}(a_j|a_i)$, $p_{dual}(a_j|q_i)$ and $p_{cross}(a_j|q_i)$ as well as $p_{dual}(q_j|a_i)$ and $p_{cross}(q_j|a_i)$. To achieve the aim of enhancement, the widely used Kullback-Leibler Divergence (KLD) is employed in this research. The loss function $\mathcal{L}_{q|q}$ defined on a mini-batch is adopted to minimize the divergence between $p_{dual}(q_j|q_i)$ and $p_{cross}(q_j|q_i)$, which can be calculated as Eq. \ref{equ:loss_qq}:
\begin{equation}\label{equ:loss_qq}
\mathcal{L}_{q|q}=\frac{1}{B}\sum^{B}_{j=1}\sum^{B}_{i=1}p_{cross}(q_{j}|q_{i})\log{\frac{p_{cross}(q_{j}|q_{i})}{p_{dual}(q_{j}|q_{i})}}
\end{equation}
where $B$ is the batch size; $i$ and $j$ are the indexes of the QA pairs in a given batch.

The same way can be used to calculate the loss functions $\mathcal{L}_{a|a}$, $\mathcal{L}_{a|q}$ and $\mathcal{L}_{q|a}$.
Then the overall loss function of GAM can be defined as Eq. \ref{equ:loss_ga}, where the hyper-parameters $\alpha_{a|q}$, $\alpha_{q|q}$, $\alpha_{q|a}$ and $\alpha_{a|a}$ are weights on different loss components:
\begin{equation}\label{equ:loss_ga}
\mathcal{L}_{ga}=\alpha_{a|q}\mathcal{L}_{a|q}+\alpha_{q|q}\mathcal{L}_{q|q}+\alpha_{q|a}\mathcal{L}_{q|a}+\alpha_{a|a}\mathcal{L}_{a|a}
\end{equation}

\subsection{Model Training and Inference}
During training stage, we jointly train the Dual-Encoders and Cross-Encoders, and align the geometry of Dual-Encoders with that of Cross-Encoders. The overall loss function to train the full model is defined as Eq. \ref{equ:overall_loss}, where $\alpha_{dual}$, $\alpha_{cross}$ and $\alpha_{ga}$ are hyper-parameters to control the loss weight.
\begin{equation}\label{equ:overall_loss}
\mathcal{L} = \alpha_{dual}\mathcal{L}_{dual} + \alpha_{cross}\mathcal{L}_{cross} + \alpha_{ga}\mathcal{L}_{ga}
\end{equation}

Since we only use the enhanced Dual-Encoders to encode questions in the inference stage while embeddings of answers are processed off-line, no extra computation is needed consequently.

\section{Experiment}
\label{sec:Experiment}

\subsection{Datasets}
\citet{DBLP:conf/acl-mrqa/AhmadCYC19} introduced the Retrieval Question-Answering (ReQA) task, which focuses on sentence-level answer retrieval and establish a pipeline to transform a reading comprehension dataset to ReQA dataset. We conduct our experiments on ReQA SQuAD and ReQA NQ established from SQuAD v1.1 \cite{2016/rajpurkar/squad} and NQ \cite{kwiatkowski-etal-2019-natural} respectively by \citet{DBLP:conf/acl-mrqa/AhmadCYC19}. We also use the same pipeline to process HotpotQA \cite{yang-etal-2018-hotpotqa} and NewsQA \cite{trischler-etal-2017-newsqa} datasets for more experiments. ReQA HotpotQA and ReQA NewsQA are used to denote the processed version of HotpotQA and NewsQA datasets respectively in this research. Since the original test sets of datasets above are not publicly available, the original validation sets are used as test sets. The statistics of ReQA datasets are shown in Table \ref{table: DatasetsStatistics}.

\subsection{Evaluation Metrics}
We adopt two popular metrics\footnote{https://github.com/google/retrieval-qa-eval} for evaluation, i.e., mean reciprocal rank (MRR) and recall at N (R@N), which are widely used for measuring retrieval-based QA task \cite{DBLP:conf/acl-mrqa/AhmadCYC19}. 

MRR is the average reciprocal ranks of retrieval results, as illustrated in Eq.~\ref{equ:mrr}, where $Q$ is a set of questions and $rank_i$ is the rank of the first correct answer for the $i^{th}$ question. 
\begin{equation}
    \label{equ:mrr}
    \mbox{MRR} = \frac{1}{|Q|}\sum_{i}^{|Q|}\frac{1}{rank_i}
\end{equation}

R@N is the recall score in top-N predicted subsets, as illustrated in Eq.~\ref{equ:R@N}, where $A_i$ is the ranked answer list for the $i^{th}$ question and $A_i^\ast$ is the corresponding correct answer set.
\begin{equation}
    \label{equ:R@N}
    \mbox{R@N} =  \frac{1}{|Q|}\sum_{i}^{|Q|}\frac{|top_N(A_i) \cap A_i^\ast|}{|A_i^\ast|}
\end{equation}

\subsection{Compared Methods}
\label{Sup: CompetitiveMethods}

\paragraph{BM25}
A classical ranking method using TF-IDF like scoring function for information retrieval \cite{DBLP:journals/ftir/RobertsonZ09}.

\paragraph{InferSent}
A universal sentence encoder trained with supervised natural language inference task, not in need of fine-tuning for specific retrieval task \cite{2017/conneau/InferSent}.

\paragraph{USE-QA}
A multi-task pre-trained model based on the Transformer, which learns universal sentence representation through a multi-feature ranking task, a translation ranking task and a natural language inference task \cite{2020/yang/USE-QA}.

\paragraph{Dual-Encoders}
The vanilla Dual-Encoders train from scratch and can be implemented using different encoders. For instance, we use Dual-BERTs to denote the Dual-Encoders using BERT as encoder.

\paragraph{Dual-VAEs}
A model trained jointly with the question-to-question and answer-to-answer reconstruction tasks using VAE \cite{2018/shen/dualVAEs}.

\paragraph{Cross-VAEs}
A model to solve one-to-many problem in answer retrieval, aligning the feature spaces of questions and answers by the question-to-answer and answer-to-question reconstruction \cite{2020/Yu/CrossVAE}.

\paragraph{ENDX-Encoders (Ours)} The
Dual-Encoders is enhanced by our ENDX framework. For instance, ENDX-BERTs is used to denote the Dual-BERTs enhanced by ENDX.

\subsection{Implementation Details}
We split the training sets of all datasets into new training set and validation set in a ratio of 9:1. The hyper-parameters are chosen according to the model performance (R@1) on validation set. Specifically, Dual-BERTs and ENDX-BERTs are initialized using BERT base model \cite{2019/devlin/BERT}, and the encoder of other models has 2 layers and uses 768-dim BERT token embedding as input. The cross attention modules of all ENDX-Encoders have 12 heads. We use AdamW optimizer \cite{loshchilov2017decoupled} to train BERT-based model with 30 epochs and linearly decay the learning rate initialized as 2e-5, and train other models with 100 epochs using constant learning rate initialized as 1e-5. We set the loss weights $\alpha_{dual}$, $\alpha_{cross}$ and $\alpha_{ga}$ to 0.25, 0.25 and 0.5 respectively. The loss weights $\alpha_{a|q}$ and $\alpha_{q|a}$ increase linearly from 0 to 0.5, while $\alpha_{q|q}$ and $\alpha_{a|a}$ increase linearly from 0 to 1e4 both over the first 5 epochs. The batch size of BERT-based model is set to 12, and that of other models is set to 100. Finally the parameters that perform best on validation set are used on test set.

\subsection{Results and Analysis}

\paragraph{Main Results}
The results on ReQA SQuAD are shown in Table \ref{table: Performance on SQuAD}. BM25 shows competitive performance, since keywords overlap is common in ReQA SQuAD. As a pre-trained universal sentence encoder without fine-tuning, InferSent does not perform well as the pre-training datasets are relatively small. USE-QA gets stronger performance because of the use of a more powerful encoder and a larger-scale pre-training dataset. Compared to Dual-VAEs, Cross-VAEs improves MRR, R@1 and R@5 by 1.32\%, 1.07\% and 2.28\% respectively, while our ENDX-BERTs outperforms the current best model Cross-VAEs \cite{2020/Yu/CrossVAE} on MRR, R@1 and R@5 by 17.88\%, 15.00\%, 21.60\% respectively and achieves new state-of-the-art result on ReQA SQuAD.

\begin{table}[h]
	\begin{center}
		\resizebox{\columnwidth}{!}
		{  
			\begin{tabular}{lccc}
				\hline
				Method & MRR & R@1 & R@5 \\
				\hline
				BM25 & 52.96 & 45.81 & 61.31 \\
				InferSent$\dagger$ & 36.90 & 27.91 & 46.92 \\
				USE-QA$\dagger$ & 61.23 & 53.16 & 69.93 \\
				Dual-VAEs$\dagger$ & 61.48 & 55.01 & 68.49 \\
				Cross-VAEs$\dagger$ & 62.29 & 55.60 & 70.05 \\
				\hline
				Dual-RNNs & 52.19 & 40.96 & 65.11 \\
				ENDX-RNNs & 53.68(\textcolor{black}{$\uparrow$}) & 42.20(\textcolor{black}{$\uparrow$}) & 67.30(\textcolor{black}{$\uparrow$}) \\
				\hline
				Dual-GRUs & 55.24 & 44.39 & 68.00 \\
				ENDX-GRUs & 58.65(\textcolor{black}{$\uparrow$}) & 48.29(\textcolor{black}{$\uparrow$}) & 70.90(\textcolor{black}{$\uparrow$}) \\
				\hline
				Dual-LSTMs & 58.77 & 49.26 & 69.79 \\
				ENDX-LSTMs & 61.00(\textcolor{black}{$\uparrow$}) & 50.79(\textcolor{black}{$\uparrow$}) & 72.87(\textcolor{black}{$\uparrow$}) \\
				\hline
				Dual-Transformers & 62.58 & 51.51 & 75.99 \\
				ENDX-Transformers & 63.73(\textcolor{black}{$\uparrow$}) & 53.41(\textcolor{black}{$\uparrow$}) & 76.02(\textcolor{black}{$\uparrow$}) \\
				\hline
				Dual-BERTs & 71.06 & 61.24 & 83.09 \\
				ENDX-BERTs & \textbf{73.43}(\textcolor{black}{$\uparrow$}) & \textbf{63.94}(\textcolor{black}{$\uparrow$}) & \textbf{85.18}(\textcolor{black}{$\uparrow$}) \\
				\hline
			\end{tabular}
		}
	\end{center}
        \caption{Performance on ReQA SQuAD dataset, where the results with $\dagger$ are reported from \cite{2020/Yu/CrossVAE}.} 
		\label{table: Performance on SQuAD}
\end{table}

\begin{table*}
	\begin{center}
		\resizebox{2\columnwidth}{!}
		{  
			\begin{tabular}{l|ccc|ccc|ccc}
				\hline
				\multirow{2}*{Method} & \multicolumn{3}{c|}{ReQA NQ} & \multicolumn{3}{c|}{ReQA HotpotQA} & \multicolumn{3}{c}{ReQA NewsQA} \\
				~ & MRR & R@1 & R@5 & MRR & R@1 & R@5 & MRR & R@1 & R@5 \\
				\hline
				Dual-RNNs & 41.45 & 26.84 & 60.01 & 22.86 & 13.83 & 32.71 & 22.23 & 12.80 & 32.64 \\
				ENDX-RNNs & 43.57(\textcolor{black}{$\uparrow$}) & 29.29(\textcolor{black}{$\uparrow$}) & 62.11(\textcolor{black}{$\uparrow$}) & 24.20(\textcolor{black}{$\uparrow$}) & 14.69(\textcolor{black}{$\uparrow$}) & 34.79(\textcolor{black}{$\uparrow$}) & 23.33(\textcolor{black}{$\uparrow$}) & 14.39(\textcolor{black}{$\uparrow$}) & 33.14(\textcolor{black}{$\uparrow$}) \\
				\hline
				Dual-GRUs & 44.99 & 31.26 & 62.31 & 25.44 & 16.61 & 34.77 & 26.21 & 16.86 & 37.16 \\
				ENDX-GRUs & 47.18(\textcolor{black}{$\uparrow$}) & 33.36(\textcolor{black}{$\uparrow$}) & 64.09(\textcolor{black}{$\uparrow$}) & 26.96(\textcolor{black}{$\uparrow$}) & 17.76(\textcolor{black}{$\uparrow$}) & 37.08(\textcolor{black}{$\uparrow$}) & 28.61(\textcolor{black}{$\uparrow$}) & 19.78(\textcolor{black}{$\uparrow$}) & 38.51(\textcolor{black}{$\uparrow$}) \\
				\hline
				Dual-LSTMs & 46.07 & 33.13 & 62.28 & 25.16 & 16.68 & 34.33 & 28.39 & 20.32 & 37.06 \\
				ENDX-LSTMs & 49.29(\textcolor{black}{$\uparrow$}) & 36.18(\textcolor{black}{$\uparrow$}) & 65.25(\textcolor{black}{$\uparrow$}) & 25.78(\textcolor{black}{$\uparrow$}) & 16.66(\textcolor{black}{$\downarrow$}) & 35.52(\textcolor{black}{$\uparrow$}) & 30.01(\textcolor{black}{$\uparrow$}) & 21.42(\textcolor{black}{$\uparrow$}) & 39.55(\textcolor{black}{$\uparrow$}) \\
				\hline
				Dual-Transformers & 46.34 & 31.90 & 64.68 & 26.22 & 15.40 & 38.64 & 27.82 & 18.00 & 38.77 \\
				ENDX-Transformers & 47.85(\textcolor{black}{$\uparrow$}) & 33.52(\textcolor{black}{$\uparrow$}) & 65.99(\textcolor{black}{$\uparrow$}) & 26.59(\textcolor{black}{$\uparrow$}) & 15.54(\textcolor{black}{$\uparrow$}) & 39.45(\textcolor{black}{$\uparrow$}) & 29.25(\textcolor{black}{$\uparrow$}) & 19.21(\textcolor{black}{$\uparrow$}) & 41.00(\textcolor{black}{$\uparrow$}) \\
				\hline
				Dual-BERTs & 54.80 & 40.58 & 72.66 & 39.04 & 27.13 & 52.91 & 37.35 & 26.64 & 49.36 \\
				ENDX-BERTs & \textbf{57.76}(\textcolor{black}{$\uparrow$}) & \textbf{43.32}(\textcolor{black}{$\uparrow$}) & \textbf{76.15}(\textcolor{black}{$\uparrow$}) & \textbf{40.68}(\textcolor{black}{$\uparrow$}) & \textbf{28.74}(\textcolor{black}{$\uparrow$}) & \textbf{54.58}(\textcolor{black}{$\uparrow$}) & \textbf{37.90}(\textcolor{black}{$\uparrow$}) & \textbf{27.26}(\textcolor{black}{$\uparrow$}) & \textbf{49.95}(\textcolor{black}{$\uparrow$}) \\
				\hline
			\end{tabular}
		}
	\end{center}
        \caption{Performance comparison on ReQA NQ, ReQA HotpotQA and ReQA NewsQA datasets.
		}
		\label{table: Performance on otherdatasets}
\end{table*}

Table \ref{table: Performance on otherdatasets} shows the performance comparison on ReQA NQ, ReQA HotpotQA and ReQA NewsQA datasets. Since the results in Table \ref{table: Performance on SQuAD} have already shown the Dual-BERTs and ENDX-BERTs can significantly outperform BM25, InferSent, USE-QA, Dual-VAEs and Cross-VAEs, we only compare Dual-Encoders and ENDX-Encoders. The results in Table \ref{table: Performance on SQuAD} and Table \ref{table: Performance on otherdatasets} both indicate the superiority of our ENDX framework which consistently outperforms Dual-Encoders with significant margins. For instance,
on MRR, R@1 and R@5, ENDX-LSTMs outperforms Dual-LSTMs by 6.99\%, 9.21\%, 4.77\% in ReQA NQ, and ENDX-Transformers outperforms Dual-Transformers by 5.14\%, 6.72\%, 5.75\% in ReQA NewsQA. 
Compared to the powerful Dual-BERTs, our ENDX-BERT shows average relative improvements over four datasets by 3.60\%, 4.86\% and 2.92\% on MRR, R@1 and R@5 respectively (t-test of 10 runs, p-values < 0.01).

\begin{table}[h]
	\begin{center}
		\resizebox{0.8\columnwidth}{!}
		{  
			\begin{tabular}{lccc}
				\hline
				Method & MRR & R@1 & R@5 \\
				\hline
				USE-QA & 47.06 & 40.90 & 53.44 \\
				Cross-VAEs & 48.52 & 44.55 & 53.52 \\
				Dual-BERTs & 60.19 & 48.56 & 74.02 \\
				ENDX-BERTs & \textbf{64.93} & \textbf{52.23} & \textbf{81.36} \\
				\hline
			\end{tabular}
		}
	\end{center}
        \caption{Performance on ReQA SQuAD sub-dataset, each answer of which has at least 8 matched questions.}
		\label{table: SubsetResults}
\end{table}

\begin{figure}[h]
\centering
\includegraphics[width=0.8\columnwidth]{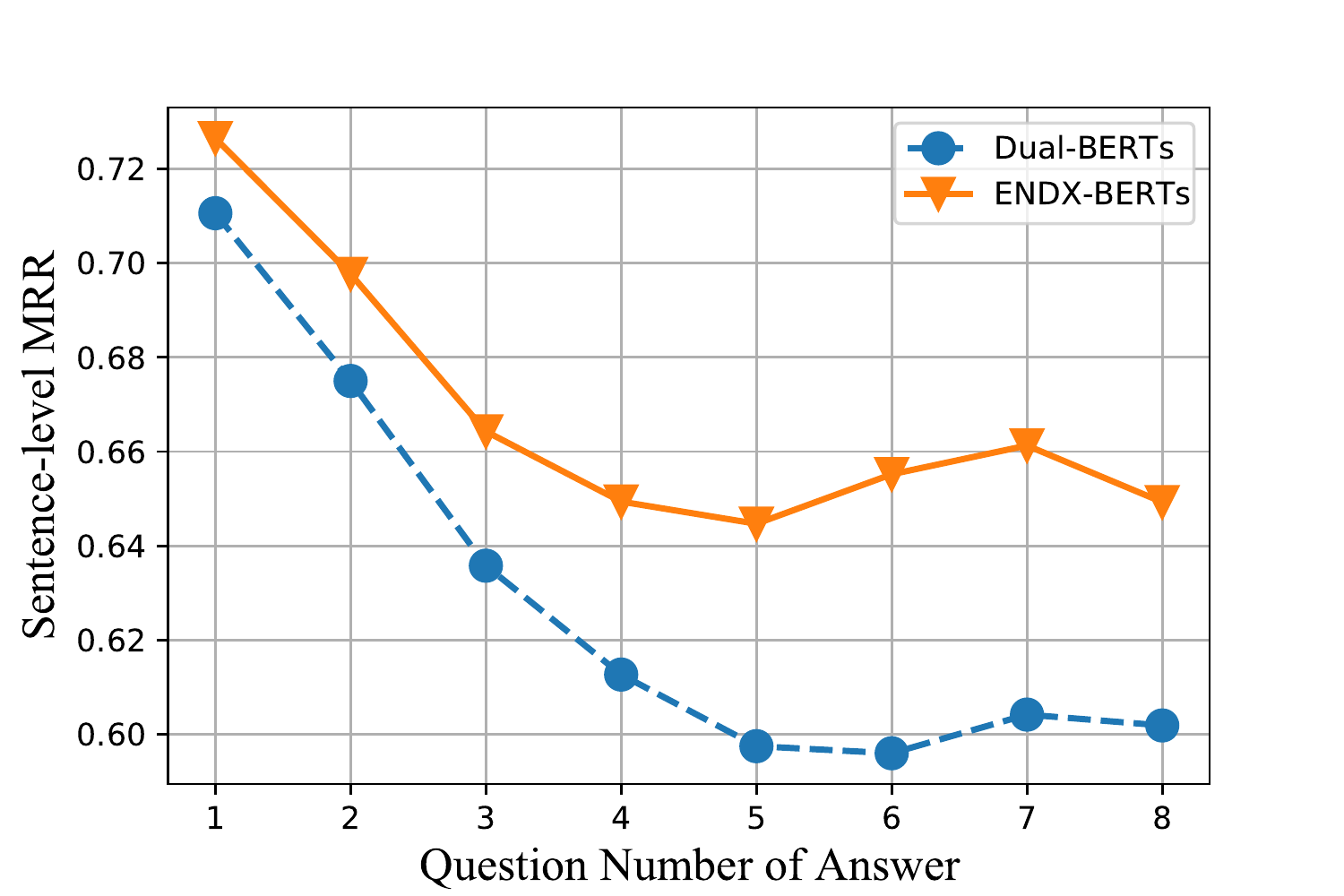}
\caption{Comparison between Dual-BERTs and ENDX-BERTs on ReQA SQuAD sub-datasets where answer has different minimum number of matched questions.}
\label{fig: SubsetResults}
\end{figure}

\paragraph{Performance on sub-dataset}
We conduct more experiments on sub-datasets of ReQA SQuAD to validate the effectiveness of our framework on coping with the one-to-many problem. The comparison results between Dual-BERTs and ENDX-BERTs on sub-datasets, in which answers have different minimum number of matched questions, are shown as Fig. \ref{fig: SubsetResults}. It is observed that ENDX-BERTs outperforms Dual-BERTs solidly. The results of the most difficult sub-dataset, in which answers have at least 8 different questions, are shown in Table \ref{table: SubsetResults}. Compared to Dual-BERTS, USE-QA and Cross-VAEs, our proposed model prominently shows the best performance under such a significant one-to-many circumstance. 

\paragraph{Analysis on the effects of GAM}
\begin{figure*}
  \centering
  \subfigure[Cross-embeddings (ENDX) ]{\label{fig:qq_post}\includegraphics[width=0.65\columnwidth]{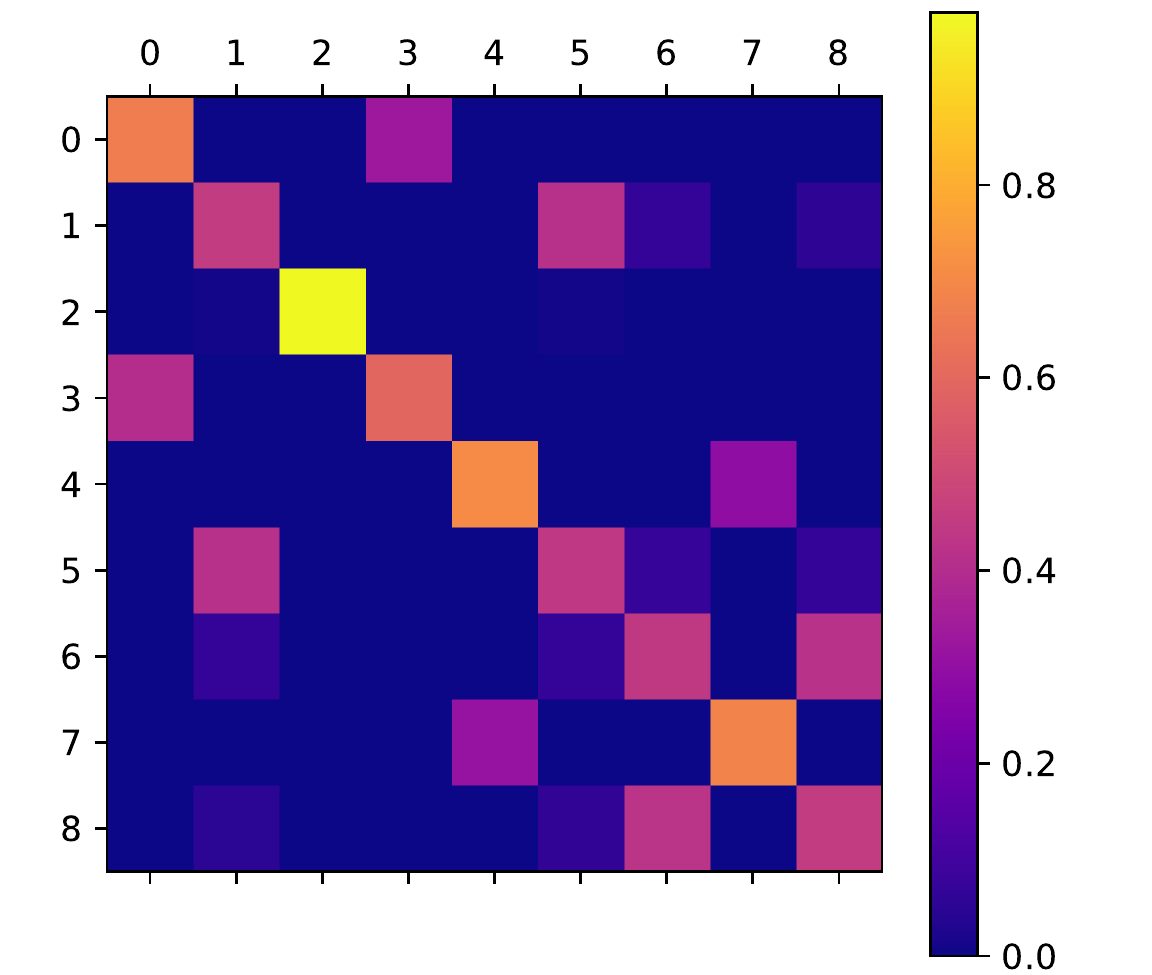}}
  \subfigure[Dual-embeddings (ENDX)]{\label{fig:qq_prior}\includegraphics[width=0.65\columnwidth]{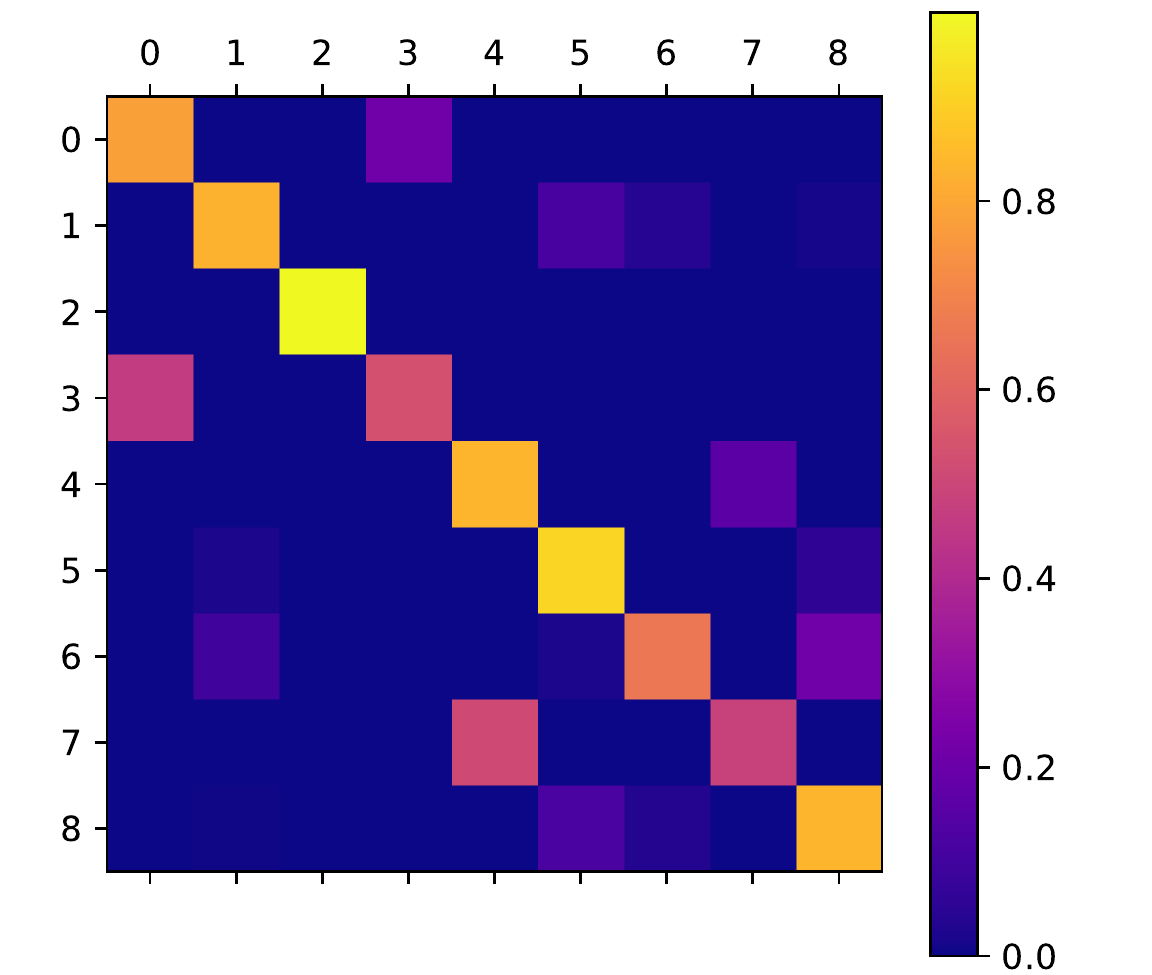}}
  \subfigure[Dual-embeddings (Baseline)]{\label{fig:qq_base}\includegraphics[width=0.65\columnwidth]{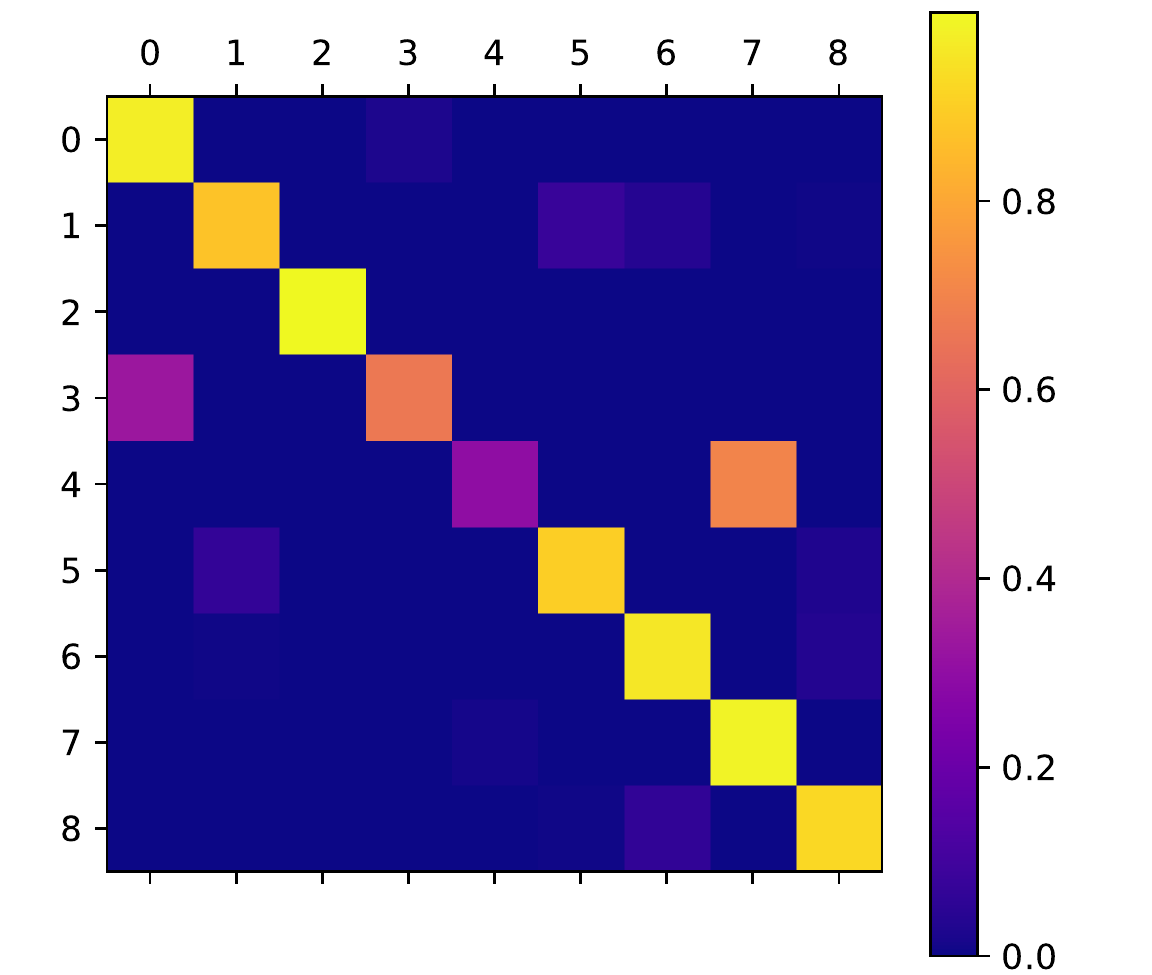}}
  \caption{Question-question similarity matrices of ENDX cross-embeddings, ENDX dual-embeddings and baseline dual-embeddings, where the $i^{th}$ row of matrix denotes the similarity between $i$th question and the others.}
  \label{fig: QQComparison}
\end{figure*} 
We also sample multiple questions with same answer and encode the questions by Cross-Encoders, Dual-Encoders enhanced with the proposed GAM, and basic Dual-Encoders, respectively. The question-question similarity matrices are visualized in Fig. \ref{fig: QQComparison}. 
In cross-embeddings, questions could attend the matched answer which results in more accurate question representations and better capture of the correlations between questions (see Fig. \ref{fig:qq_post}). During ENDX training, we use GAM to align the geometry of dual-embeddings with that of cross-embeddings. As shown in Fig. \ref{fig:qq_prior}, dual-embeddings enhanced by GAM are able to capture more correlations in question-question similarities compared to baseline dual-embeddings (Fig. \ref{fig:qq_base}). 

\paragraph{Ablation study on loss function of GAM}
We perform the ablation study on the proposed ENDX-BERTs in ReQA NQ by removing different components of GAM loss function. As shown in Table \ref{table: AblationOnNQ}, all metric scores drop significantly without optimizing $\mathcal{L}_{a|q}$ or $\mathcal{L}_{q|a}$, which indicates that $p(a_{j}|q_{i})$ and $p(q_{j}|a_{i})$ describe the most important parts of geometry. The reason we conjecture is that the answer retrieval task focus more on the relative distance of question-to-answer in feature space, while $\mathcal{L}_{q|q}$ and $\mathcal{L}_{a|a}$ are also helpful.

\begin{table}[h]
	\begin{center}
		\resizebox{0.8\columnwidth}{!}
		{  
			\begin{tabular}{lccc}
				\hline
				Method & MRR & R@1 & R@5 \\
				\hline
				Dual-BERTs & 54.80 & 40.58 & 72.66 \\
				ENDX-BERTs & \textbf{57.76} & \textbf{43.32} & \textbf{76.15} \\
				\quad \quad \quad w/o $\mathcal{L}_{q|q}$ & 56.37 & 42.01 & 74.65 \\
				\quad \quad \quad w/o $\mathcal{L}_{a|a}$ & 56.89 & 42.25 & 75.28 \\
				\quad \quad \quad w/o $\mathcal{L}_{q|a}$ & 56.00 & 42.12 & 73.27 \\
				\quad \quad \quad w/o $\mathcal{L}_{a|q}$ & 55.87 & 41.39 & 74.14 \\
				\hline
			\end{tabular}
		}
	\end{center}
        \caption{Ablation study on ReQA NQ dataset.}
		\label{table: AblationOnNQ}
\end{table}

\paragraph{Comparison with BERT$\rm{_{QA}}$}
We also compare the proposed ENDX-BERTs against the interaction-based model $\rm{BERT_{QA}}$ \cite{2019/devlin/BERT}, which encodes concatenated sequence for every candidate QA pair. Due to the extremely large computational cost of $\rm{BERT_{QA}}$, we only sample 500 QA pairs from 27 passages in ReQA SQuAD as the test set. The experimental result is shown in Table \ref{table: SingleTower}, where ENDX-BERTs improves MRR, R@1 and R@5 over Dual-BERTs by +4.61\%, +4.59\% and +5.44\% respectively and only falls behind $\rm{BERT_{QA}}$ by -3.38\%, -4.93\% and -0.81\%. However, the inference runtime complexity is significantly reduced from $O(n \times m)$ to $O(n+m)$ compared to $\rm{BERT_{QA}}$, where $n$ and $m$ are the numbers of questions and answers respectively. Therefore, the propopsed ENDX-BERTs can better balance between accuracy and efficiency for answer retrieval.
\begin{table}[h]
		
	\begin{center}
		\resizebox{\columnwidth}{!}
		{  
			\begin{tabular}{lcccc}
				\hline
				Method & MRR & R@1 & R@5 & average ms \\
				\hline
				Dual-BERTs & 71.83 & 60.42 & 87.26 & 14.19\\
				ENDX-BERTs & 76.44 & 65.01 & 92.70 & 14.19 \\
				BERT$_{\rm QA}$ & 79.82 & 69.94 & 93.51 & 6939.61 \\
				\hline
			\end{tabular}
		}
	\end{center}
	\caption{Comparison with BERT$_{\rm QA}$, where the average time (ms) to retrieve answer for one question is tested on one NVIDIA Tesla V100 GPU.}
		\label{table: SingleTower}
\end{table}

\subsection{Case Study}
\label{Sup: CaseStudy}
\begin{figure}[h]
\centering
\includegraphics[width=0.8\columnwidth]{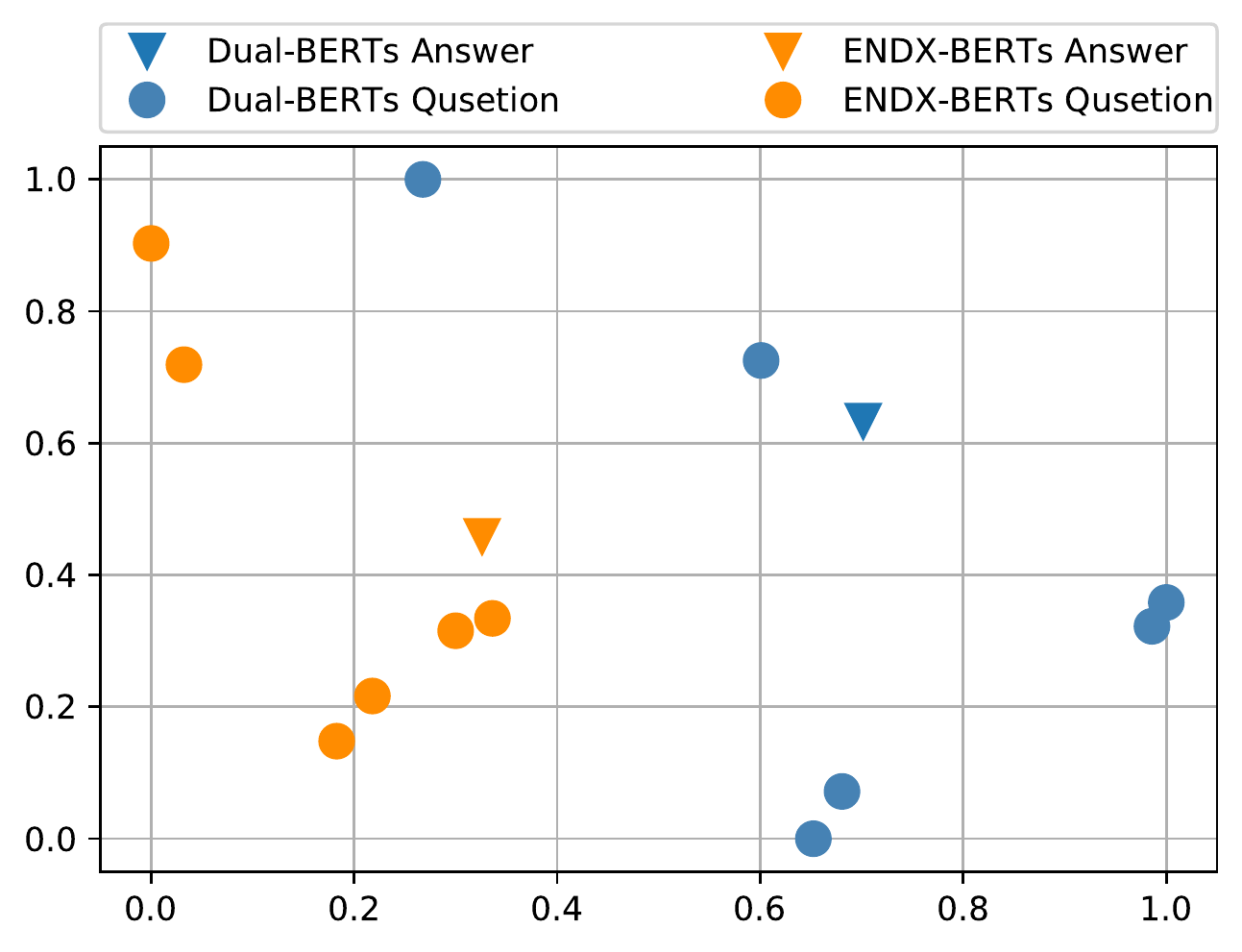}
\caption{A case of 6 different questions sharing one answer, where the blue dot and yellow dot present the question and answer embeddings of Dual-BERTs and ENDX-BERTs in 2D space respectively.}
\label{fig: QAScatter}
\end{figure}

Figure \ref{fig: QAScatter} shows the dual-embeddings projection (t-SNE, \citealp{van2008visualizing}) of 6 different questions and their shared answer. It can be seen that the dual-embeddings of our ENDX-BERTs are more compact than that of Dual-BERTs, which proves that our method can better align the questions and answers, and can produce more general representation to alleviate the one-to-many problem. 

\section{CONCLUSION}
In this work, we propose a framework that enhances Dual-Encoders with cross-embeddings for answer retrieval. A novel geometry alignment mechanism is introduced to align the geometry of Dual-Encoders with cross-embeddings. Extensive experimental results show that our method significantly improves Dual-Encoders model and outperforms the state-of-the-art method on multiple answer retrieval datasets.


\bibliography{emnlp2021}
\bibliographystyle{acl_natbib}

\end{document}